\documentclass[runningheads]{llncs}
\usepackage{graphicx}
\usepackage{amsmath,amssymb,amsfonts}
\usepackage{multirow}
\usepackage{changes}
\usepackage[hidelinks]{hyperref}
\begin{document}
\title{Polar Transformation Based Multiple Instance Learning Assisting Weakly Supervised Image Segmentation With Loose Bounding Box Annotations}
%
%
\author{Juan Wang\inst{1}\orcidID{0000-0003-3124-9901} \and
Bin Xia\inst{2}\orcidID{0000-0002-0340-8082}}
\authorrunning{J. Wang and B. Xia}
%
\institute{Delta Micro Technology, Inc., Laguna Hills, CA 92653 USA 
\email{wangjuan313@gmail.com}\\
\and
Shenzhen SiBright Co. Ltd., Shenzhen, Guangdong 518052 China\\
\email{b.xia@sibionics.com}}
\maketitle              
\begin{abstract}
This study investigates weakly supervised image segmentation using loose bounding box supervision. It presents a multiple instance learning strategy based on polar transformation to assist image segmentation when loose bounding boxes are employed as supervision. In this strategy, weighted smooth maximum approximation is introduced to 
incorporate the observation that pixels closer to the origin of the polar transformation are more likely to belong to the object in the bounding box. The proposed approach was evaluated on a pubic medical dataset using Dice coefficient. The results demonstrate its superior performance. The codes are available at \url{https://github.com/wangjuan313/wsis-polartransform}.

\keywords{Polar transformation \and Loose bounding box \and Multiple instance learning \and Weakly supervised image segmentation \and Deep neural networks.}
\end{abstract}

\section{Introduction}
Image segmentation is the process of assigning a category label to every pixel in an image such that pixels with the same label share certain characteristics. In recent years, with the success of the deep learning in medical image analysis \cite{wang2017detecting,esteva2017dermatologist,wang2018context,wang2020simultaneous}, deep neural networks (DNNs) have been used to tackle a variety of image segmentation tasks in a fully-supervised manner \cite{ronneberger2015u,long2015fully,chen2018encoder}. However, collecting large-scale dataset with precise pixel-wise annotation for DNN training is labor-intensive and expensive, thus limiting the value of the image segmentation in real applications, which is especially true in medical imaging. 

To tackle this issue, great efforts have been made to develop weakly supervised image segmentation (WSIS) using all kinds of supervision. Among them, bounding box supervision is especially interesting. For example, Rajchl \textit{et al.} \cite{rajchl2016deepcut} trained a neural network classifier using bounding box annotations for image segmentation in an iterative optimization way. Hsu \textit{et al.} \cite{hsu2019weakly} considered mask R-CNN for simultaneous object detection and image segmentation, in which the bounding box supervision was formulated as multiple instance learning (MIL). Kervadec \textit{et al.} \cite{kervadec2020bounding} imposed a set of constraints on the network outputs based on the tightness prior of bounding boxes for image segmentation.

Recently, a generalized MIL approach \cite{wang2021bounding} was developed by considering tight bounding boxes as supervision for image segmentation and achieved state-of-the-art performance \cite{wang2021bounding,wang2021accurate}. Building on the previous success in \cite{wang2021bounding}, this work investigates the use of \textit{loose} bounding boxes as supervision to assist image segmentation. Compared with tight bounding box supervision used in \cite{wang2021bounding}, loose bounding box supervision alleviates the difficulty in obtaining annotations, thus is more beneficial in real applications. For this purpose, we propose a MIL strategy based on polar transformation of the image region in the bounding box, in which weighted smooth maximum approximation is exploited to incorporate the observation that pixels closer to the origin of the polar transformation are more likely to belong to the object in the bounding box. In this study, the proposed MIL strategy is used to assist the generalized MIL in \cite{wang2021bounding} for image segmentation. The experiments show the superior performance of the proposed approach. 

\section{Methods}

\subsection{Problem descriptions}
This study considers deep neural networks for weakly supervised image segmentation by employing loose bounding box supervision, wherein a network is employed to determine whether each pixel in the input image belongs to a category or not. Let $I$ be the input image, $Y \in \{1,2,\cdots,C\}$ is its corresponding pixel-level category label for $C$ categories under consideration, and $B = \{b_m, y_m\}, m = 1, 2, \cdots, M$ is its bounding box label with $M$ bounding box annotations, where the location label $b_m$ is a 4-dimensional vector denoting the top left and bottom right points of the bounding box, and $y_m \in \{1,2,\cdots,C\}$ is the category label of the object in the bounding box. For a given set of $N$ training images $\{(I_n, B_n), n=1,2,\cdots,N\}$, we first train a network model, and subsequently apply it to obtain the prediction of any unseen images. 

\subsection{MIL for bounding box annotation} \label{section:mil_bd}
The bounding box of an object indicates that the location label of the bounding box is the rectangle enclosing the whole object, thus the object much inside the bounding box, and does not overlap with the region outside the bounding box. In this study, line-of-interests (LoIs) of a bounding box are defined as any lines with one endpoint (denoted as point $O$) located on a pixel belonging to the object in the bounding box and the other endpoint located in the four sides of the bounding box. Therefore, for an object with category $c$ in an image $I$, any LoI of its bounding box has at least one pixel belonging to category $c$, and any pixels outside of any bounding boxes of category $c$ do not belong to category $c$. Note these observations are valid for both tight and loose bounding boxes. Based on these observations, the positive and negative bags are defined as follows:

\textit{Positive bags:} For an object with category $c$, pixels in a LoI of its bounding box compose a positive bag for category $c$. At a given point $O$, multiple LoIs can be obtained, yielding multiple positive bags for the object. As examples, in Fig.~\ref{fig:mil_demonstration}, we show positive bags of an object (i.e. sheep), in which the bounding box is denoted as red rectangle, the point $O$ is marked by green dot, and examples of positive bags are indicated by blue hashed lines. 

\textit{Negative bags:} For a category $c$, a negative bag constitutes of an individual pixel outside of any bounding boxes of category $c$. Hence, if $M$ pixels are outside of any bounding boxes of category $c$ in an image, then $M$ negative bags are generated.

\begin{figure}[htbp] 
	\centering
	\includegraphics[trim=0in 0in 0in 0in,clip,width=1.8in]{000000046804_bb.png}
	\caption{Demonstration of positive bags in bounding box annotation. In this plot, the bounding box of the object is marked by the red rectangle, the point $O$ is denoted by the green dot, and examples of positive bags are indicated by blue dashed lines.}
	\label{fig:mil_demonstration}
\end{figure}

\subsection{Polar transformation} \label{section:pt}
Note LoIs of any bounding box can be obtained by applying polar transformation to the image region in  the bounding box. The polar transformation of an image transfers the image from the Cartesian coordinate system to the polar coordinate system, providing a pixel-wise representation in the polar coordinate system. 

Suppose $(u,v)$ is the Cartesian coordinate of a pixel in the image with respect to the origin (which is the point $O$ in Section \ref{section:mil_bd} in this study), and its corresponding polar coordinate is $(r, \theta)$, where $r > 0$ and $\theta \in [0, 2\pi]$ are the radial and angular coordinates, respectively. The polar transformation maps the pixel $(u,v)$ in the Cartesian coordinate plane to the corresponding pixel $(r, \theta)$ in the polar coordinate plane as follows: 
\begin{equation}
\begin{array}{c}
r = \sqrt{u^2+v^2} \\
\theta = \tan^{-1}(v/u) \\
\end{array}
\label{equ:polar_transformation}
\end{equation}
With polar transformation, a LoI of the image region in the bounding box is converted into a horizontal line in its transferred polar image. 

In polar transformation, one needs to preset the output shape of the transferred polar image (denoted as $N_r \times N_{\theta}$) and the radius $R$ of the transformation. In the end, the radial coordinate $r$ is evenly distributed in $[0, R]$ with step $R / N_r$, and the angular coordinate $\theta$ is evenly distributed in $[0, 2\pi]$ with step $2\pi / N_{\theta}$.

\begin{figure}[htbp] 
	\centering
	\setlength{\tabcolsep}{2pt}
	\begin{tabular}{ccc}
	\includegraphics[trim=0in 0in 0in 0in,clip,height=1.5in]{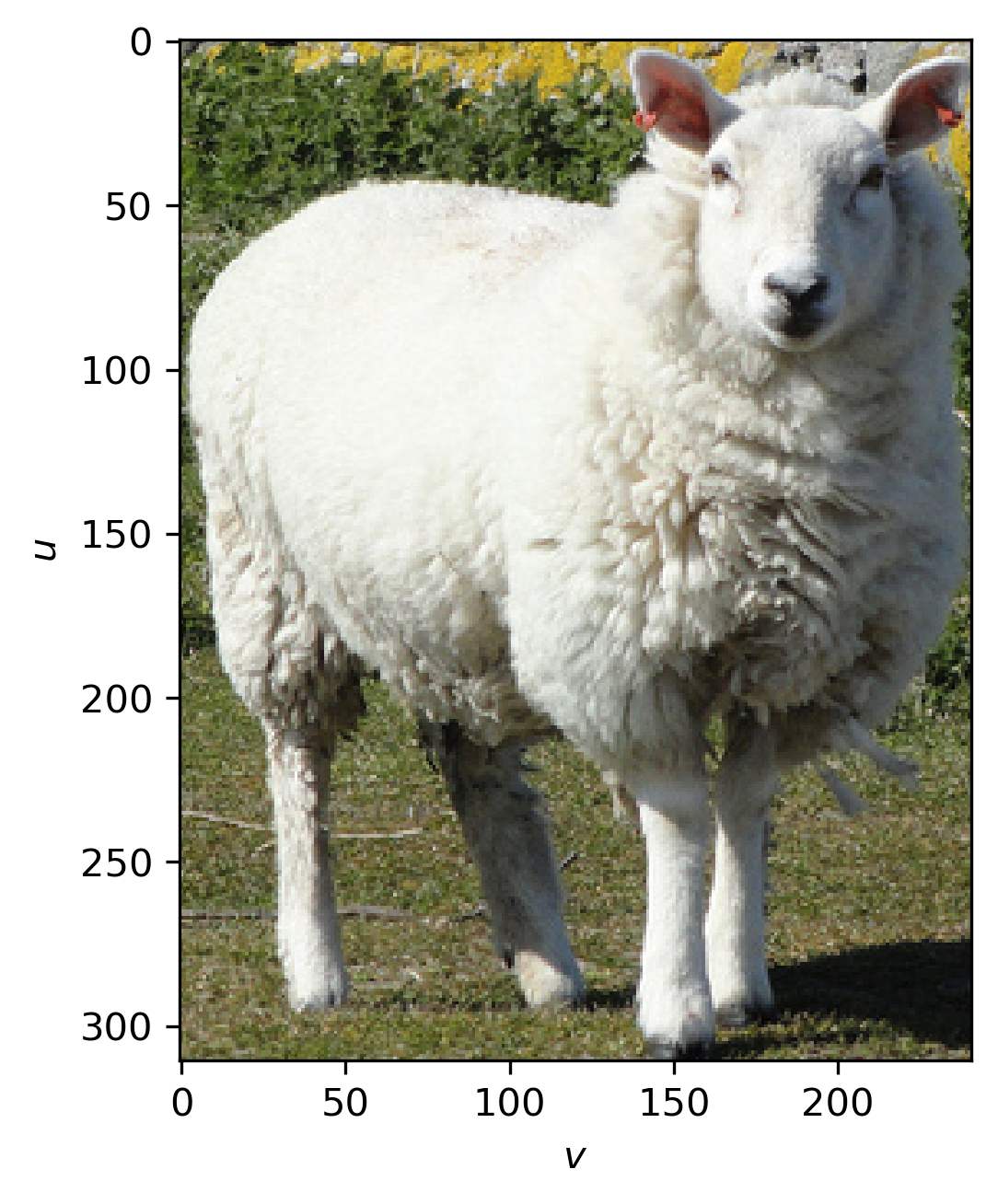} &
	\includegraphics[trim=0in 0in 0in 0in,clip,height=1.5in]{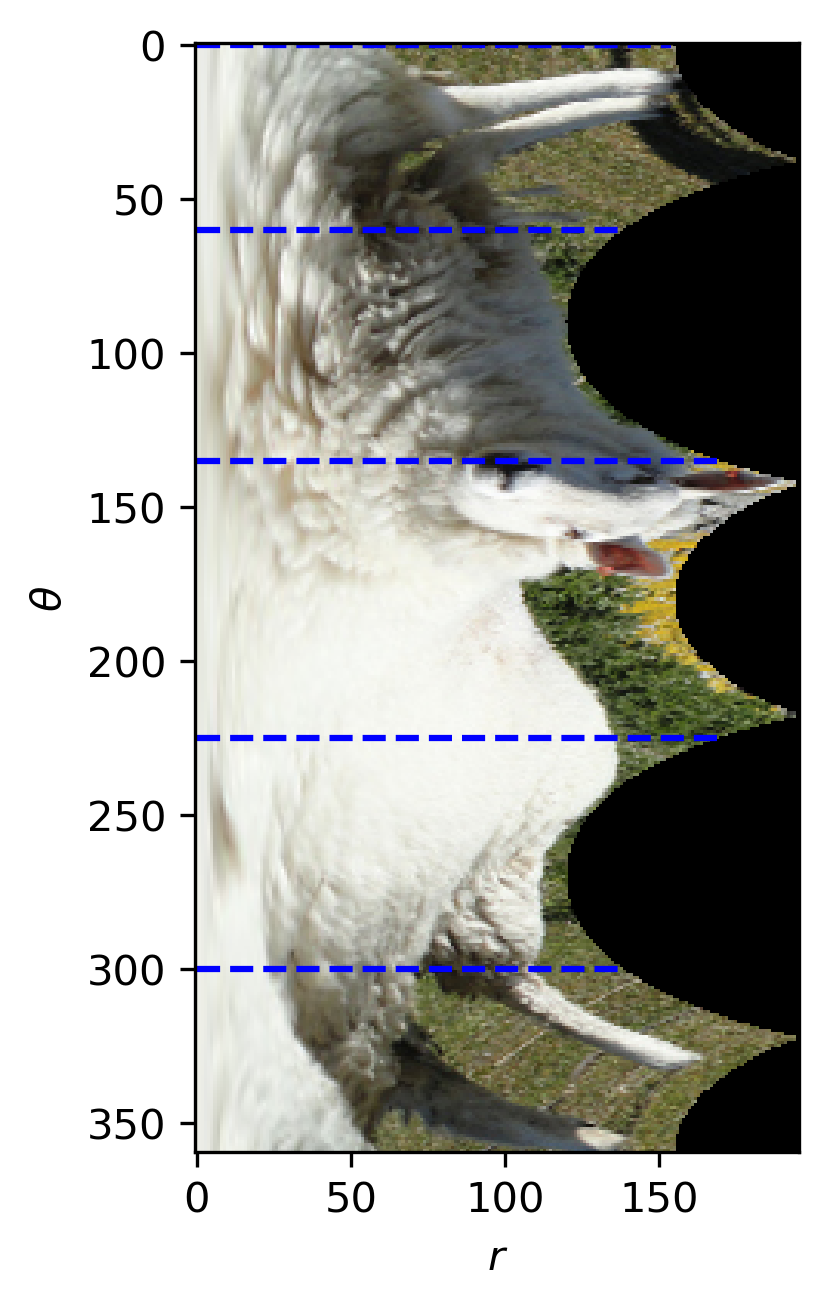} & 				    \includegraphics[trim=0in 0in 0in 0in,clip,height=1.5in]{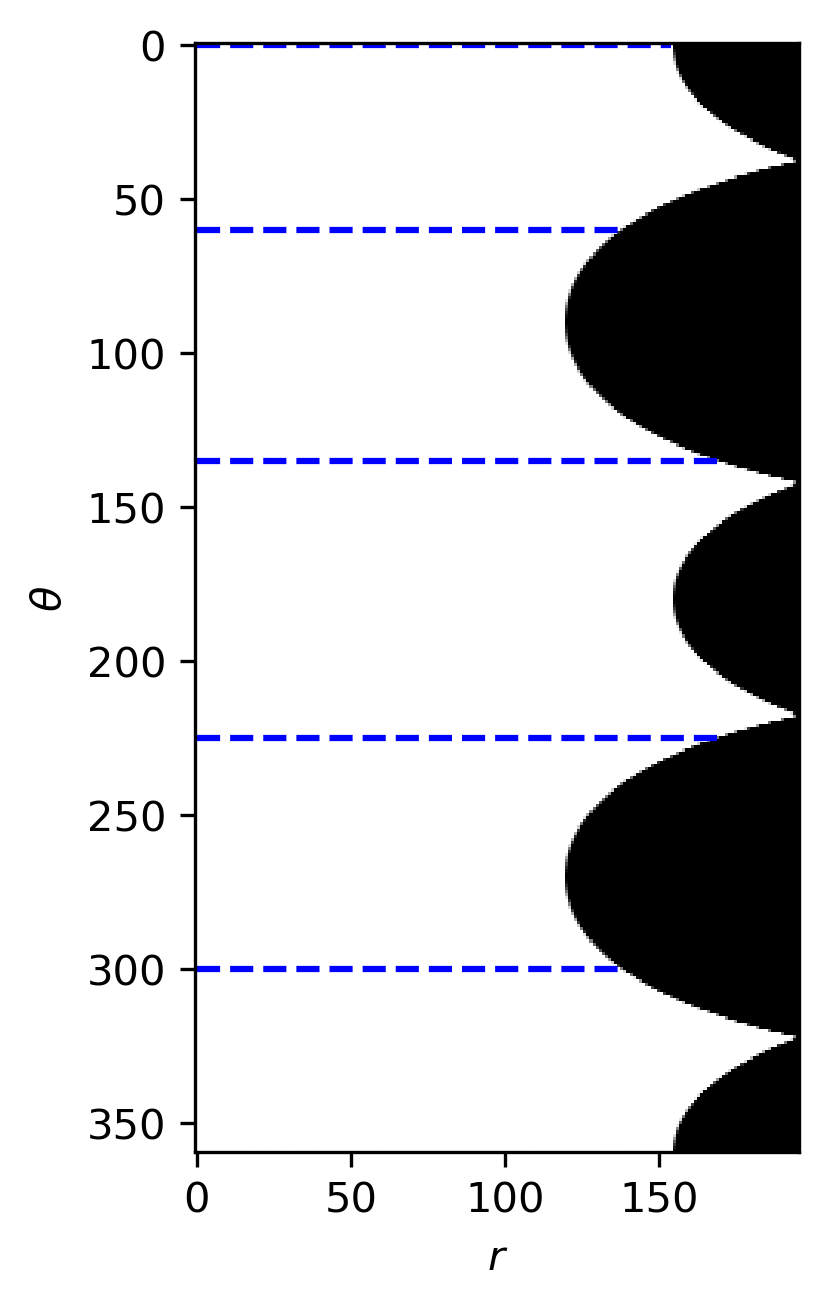} \\ 
	(a) & (b) & (c) \\
	\end{tabular}
	\caption{Demonstration of polar transformation. (a) The image region in the bounding box as shown in Fig. \ref{fig:mil_demonstration}. (b) The transferred polar image of (a), in which LoIs in Fig.~\ref{fig:mil_demonstration} are converted into the horizontal dashed lines marked by blue color. (c) The transferred polar image of the corresponding binary bounding box region. }
	\label{fig:polar_transformation}
\end{figure}

In Fig.~\ref{fig:polar_transformation}, we demonstrate an example of polar transformation of an image region. Fig.~\ref{fig:polar_transformation}(a) shows the image region in the bounding box in Fig.~\ref{fig:mil_demonstration}. Its transferred polar image is shown in Fig.~\ref{fig:polar_transformation}(b), in which blue dashed lines correspond to LoIs marked in Fig.~\ref{fig:mil_demonstration}. During polar transformation, the following parameters are used: $O$ is the center of the bounding box, $N_r$ and $R$ are the half length of the diagonal line of the bounding box, and $N_{\theta}=360$. 


\subsection{Positive bag prediction calculation}
A positive bag contains at least one pixel in the object, hence the pixel with highest prediction tends to be in the object. Therefore, the prediction of the bag $b$ being positive for category $c$ is $P_c(b) = \max_{k=0}^{n-1} p_{kc}$, where $n$ is the number of pixels in the bag $b$, $p_{kc}$ is the network output of the pixel location $k$ along the radial coordinate for category $c$ in the transferred polar image. In $p_{kc}$, $k=0$ denotes the pixel of the origin $O$ and $k=n-1$ is the pixel located in the four sides of the bounding box.

Note the polar transformation of the rectangle image region leads to variable $n$ and $n < N_r$ in the transferred polar image as shown in Fig.~\ref{fig:polar_transformation}(b). To determine $n$ for LoIs, we apply the same polar transformation to the binary bounding box region, and then employ its values to determine $n$. As example, Fig.~\ref{fig:polar_transformation}(c) shows the transferred polar image of the binary bounding box region of the image in Fig.~\ref{fig:polar_transformation}(a), where the pixels with white color are in LoIs.

Finally, the origin $O$ is determined based on the condition that it is inside the object in the bounding box. It is selected as the pixel with maximum network output among all of the pixels in the bounding box during training. Such design is intuitive since the pixel with highest prediction are more likely be in the object.

\subsection{Loss function} 
Suppose positive and negative bags of category $c$ are $\mathcal{B}_c^+$ and $\mathcal{B}_c^-$, respectively, then the loss $\mathcal{L}_c$ for polar transformation based MIL are as follows:
\begin{equation}
\mathcal{L}_c = \phi_c(P; \mathcal{B}_c^+, \mathcal{B}_c^-) + \lambda \varphi_c(P)
\label{equ:mil_loss}
\end{equation}
where $\phi_c$ is the unary loss, $\varphi_c$ is the pairwise loss, and $\lambda$ is a constant value controlling the trade off between the two losses.

Due to the imbalance between positive and negative bags, the unary loss $\phi_c$ is defined as focal loss \cite{ross2017focal} for bag prediction:
\begin{equation}
\phi_c = -\frac{1}{N^+} \left( \sum_{b \in \mathcal{B}_c^+} \beta \left(1-P_c(b)\right)^\gamma \log P_c(b) +  \sum_{b \in \mathcal{B}_c^-} (1-\beta)P_c(b)^\gamma \log(1-P_c(b)) \right)
\label{equ:unary_loss}
\end{equation}
where $N^+ = \max(1, |\mathcal{B}_c^+|)$, $\beta \in [0,1]$ is the weighting factor, and $\gamma \geq 0$ is the focusing parameter.

The pairwise loss is used to impose the piece-wise smoothness on the network output as follows:
\begin{equation}
\varphi_c = \frac{1}{|\varepsilon|} \sum_{(k,k^\prime) \in \varepsilon} \left( q_{kc} - q_{k^\prime c} \right) ^2
\end{equation}
where $q_{kc}$ is the network output of the pixel location \textit{k} for category $c$ in the image domain and $\varepsilon$ is the set containing all neighboring pixel pairs. 

For all $C$ categories, the loss $\mathcal{L}_{p}$ of the polar transformation based MIL is:
\begin{equation}
\mathcal{L}_{p} = \sum_{c=1}^C \mathcal{L}_c
\end{equation}

Finally, as noted in the introduction, the proposed polar transformation based MIL is used to assist the generalized MIL in \cite{wang2021bounding}. Suppose the generalized MIL loss is $\mathcal{L}_{g}$, the loss used in this study for network optimization is:
\begin{equation}
\mathcal{L} = \mathcal{L}_{p} + \mathcal{L}_{g}
\end{equation}

\subsection{Weighted smooth maximum approximation}
In this study, the prediction of the bag $b$ being positive for category $c$ is $P_c(b) = \max_{k=0}^{n-1} p_{kc}$, hence its derivative $\partial P_c / \partial p_{kc}$ is discontinuous, leading to numerical instability. To deal with this issue, smooth maximum approximation is considered as in  \cite{wang2021bounding}. Moreover, in LoI the pixels closer to the origin $O$ are more likely belonging to the object. To incorporate this observation, a weight is introduced in the smooth maximum approximation. In this study, two variants of weighted smooth maximum approximation as follows are considered. 

(1) \textit{weighted $\alpha$-softmax function:}
\begin{equation}
S_{\alpha}(b) = \frac{\sum_{k=0}^{n-1} w_kp_{kc} e^{\alpha w_kp_{kc}}}{\sum_{k=0}^{n-1} e^{\alpha w_kp_{kc}}}
\label{equ:softmax_approx}
\end{equation}

(2) \textit{weighted $\alpha$-quasimax function:}
\begin{equation}
Q_{\alpha}(b) = \frac{1}{\alpha} \log \left(\sum_{k=0}^{n-1} e^{\alpha w_kp_{kc}}\right) - \frac{\log n}{\alpha}
\label{equ:quasimax_approx}
\end{equation}
In these two equations, $\alpha>0$ is a constant and $0 \leq w_k \leq 1$ is the weight of $p_{kc}$. 

To explicitly incorporate the observation mentioned above, the weight $w_k$ is defined as:
\begin{equation}
w_k = e^{-k^2/(2\sigma^2)}
\end{equation}
where $\sigma = (N_r-1)/\sqrt{-2\log w_{min}}$ and $w_{min}$ is a preset parameter for the minimum weight of the pixel in the bag. In the transferred polar image, the minimum weight is given to the pixel furthest from the origin $O$, which corresponds to the pixel in the four sides of the bounding box in the image domain.

\section{Experiments}

\subsection{Dataset}
This study made use of the prostate MR image segmentation 2012 dataset for performance evaluation. It was developed in MICCAI 2012 grand challenge \cite{litjens2014evaluation}, including both benign and malignant cases. The images in the dataset are the transversal T2-weighted MR images, which were acquired at different centers with multiple MRI vendors and different scanning protocols. In this study, the dataset was divided into two non-overlapping subsets as in \cite{wang2021bounding}, one with 40 patients for training and the other with 10 patients for validation. 

\subsection{Implementation details}
This study implemented the experiments using PyTorch. Image segmentation was conducted on the 2D slices of MR images. The loose bounding boxes were obtained by adding margin of 5 pixels on each side of the tight bounding boxes, which were converted from the corresponding segmentation masks available in the dataset. The parameters in the loss $\mathcal{L}_{p}$ were set as $\lambda=10$ (equation \eqref{equ:mil_loss}) based on experience, and $\beta=0.25$ and $\gamma=2$ (equation \eqref{equ:unary_loss}) according to the focal loss \cite{ross2017focal}. The parameters for polar transformation were set as $N_r=R=30$ and $N_{\theta}=90$ based on experience. The parameters $\alpha$ and $w_{min}$ in weighted smooth maximum approximation were obtained by grid search. 

For fairness of comparison, the other experimental setups were same as those in study \cite{wang2021bounding}. That is, the network for image segmentation was a residual version of UNet \cite{ronneberger2015u}. The Adam optimizer \cite{kingma2014adam} were used for model training and its parameters are set as: batch size = 16, initial learning rate = $10^{-4}$, $\beta_1 = 0.9$, and $\beta_2 = 0.99$. The following off-line data augmentation procedure was applied to the images in the training subset: 1) mirroring, 2) flipping, and 3) rotation.

\subsection{Performance evaluation}
To evaluate the performance of the proposed approach, the Dice coefficient was considered, which has been widely used in medical image segmentation. In this study, the Dice coefficient was calculated based on 3D MR images by stacking predictions of the corresponding 2D slices together.

In the experiments, we considered the tight bounding box supervision as baseline, in which the generalized MIL approach \cite{wang2021bounding} was used. For performance comparison in the loose bounding box supervision setting, the following two approaches were employed: 1) the generalized MIL \cite{wang2021bounding} (optimized by the loss $\mathcal{L}_{g}$) and 2) the polar transformation based MIL (optimized by the loss $\mathcal{L}_{p}$). 

\section{Results}

\subsection{Main results}

Table \ref{table:promise_results} gives Dice coefficients of the proposed approach using the loose bounding box supervision. The value is 0.880 for weighted $\alpha$-softmax approximation and 0.876 for weighted $\alpha$-quasimax approximation. In comparison, the generalized MIL gets Dice coefficient of 0.859 for $\alpha$-softmax approximation and 0.866 for $\alpha$-quasimax approximation, lower than those from the proposed approach. Moreover, the polar transformation based MIL achieves worse Dice coefficients. 

\renewcommand\arraystretch{1.3}
\begin{table*}
\caption{Comparison of Dice coefficients for different methods.}
\centering
\setlength{\tabcolsep}{5pt}
\begin{tabular}{cccc}
\hline
\hline
Method & (weighted) $\alpha$-softmax & (weighted) $\alpha$-quasimax \\
\hline
Baseline & 0.878 (0.031) & 0.880 (0.024) \\
Generalized MIL & 0.859 (0.044) & 0.866 (0.033) \\
Polar transformation based MIL & 0.852 (0.024) & 0.853 (0.036)\\
Proposed approach & \textbf{0.880 (0.027)} & \textbf{0.876 (0.026)} \\
\hline
\hline
\end{tabular}
\label{table:promise_results}
\end{table*}

Furthermore, we also report results of the baseline which uses tight bounding box supervision in Table \ref{table:promise_results}. It gets Dice coefficient of 0.878 for $\alpha$-softmax approximation and 0.880 for $\alpha$-quasimax approximation, almost same as those from the proposed approach. Lastly, as the upper bound of segmentation performance, the fully supervised image segmentation gets Dice coefficient of 0.894.

\subsection{Performance sensitivity to $\alpha$ and $w_{min}$}
\begin{figure}[htbp] 
	\centering
	\setlength{\tabcolsep}{6pt}
	\begin{tabular}{cc}
	\includegraphics[trim=0in 0in 0in 0in,clip,width=1.7in]{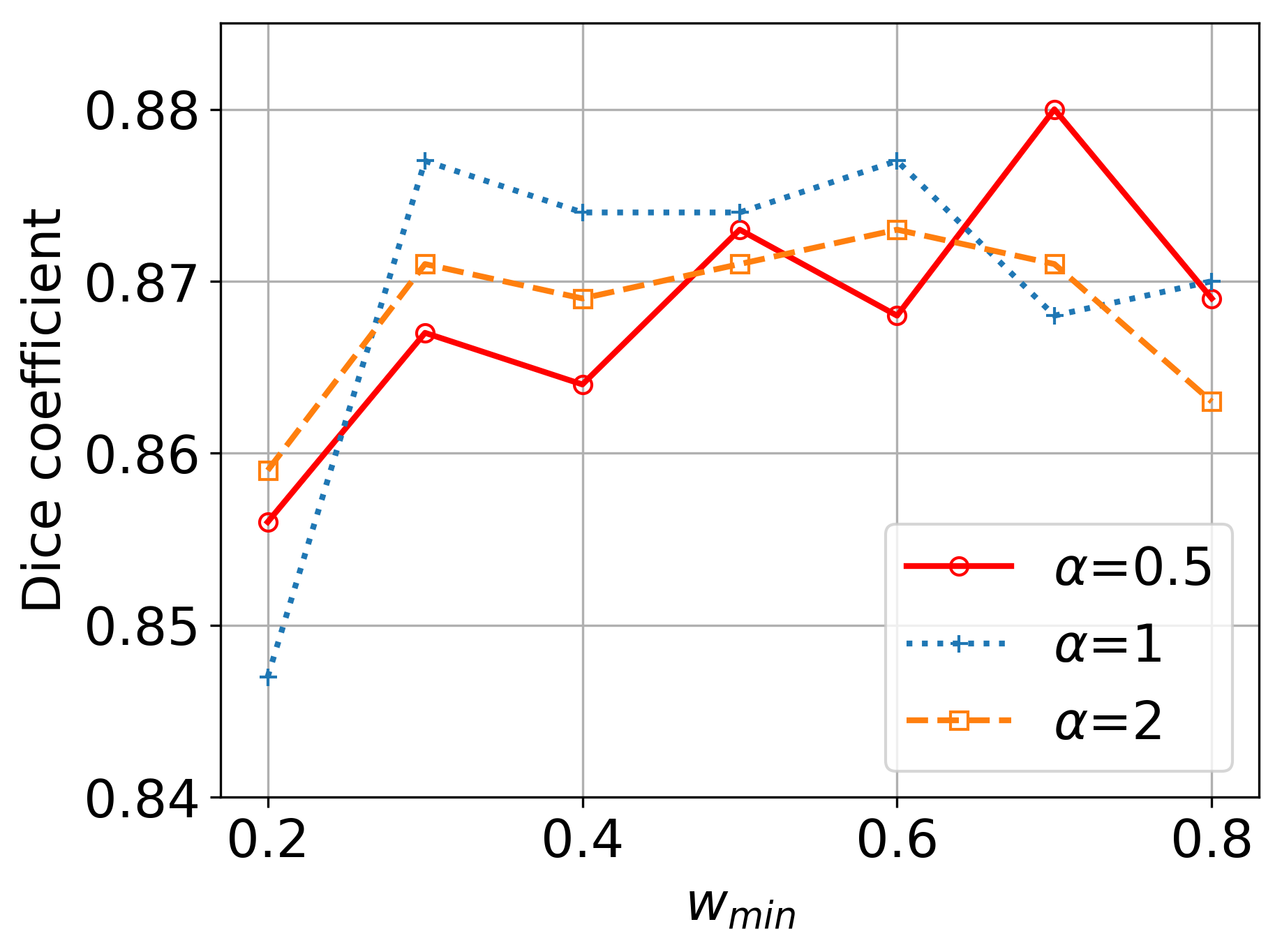} &
	\includegraphics[trim=0in 0in 0in 0in,clip,width=1.7in]{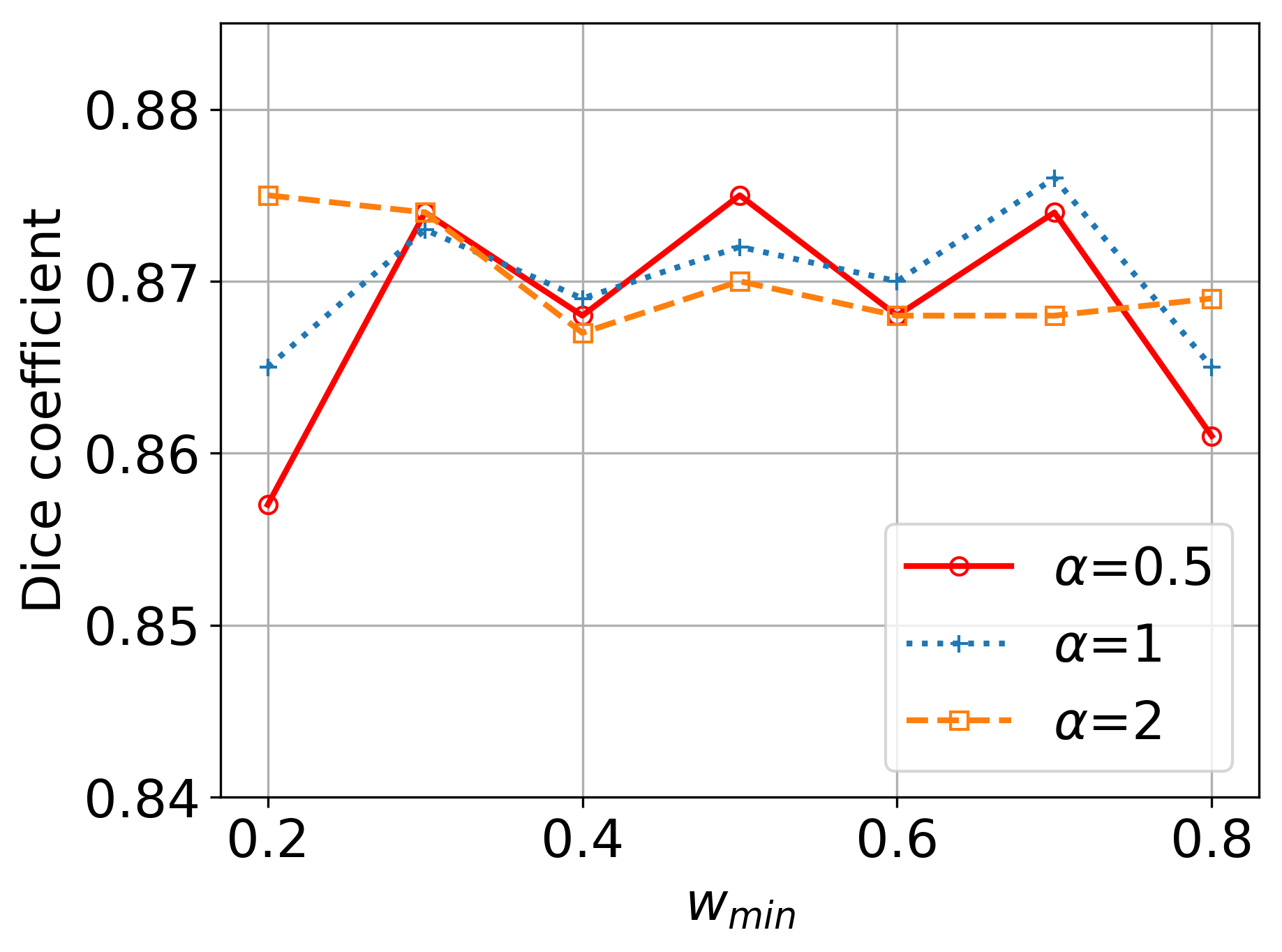} \\ 
	(a) weighted $\alpha$-softmax & (b) weighted $\alpha$-quasimax \\
	\end{tabular}
	\caption{Dice coefficients of the proposed approach on different $\alpha$'s and $w_{min}$'s.}
	\label{fig:sensitivity_perf}
\end{figure}

To evaluate sensitivity of the proposed approach to parameters in weighted smooth maximum approximation, Fig.~\ref{fig:sensitivity_perf}(a) shows Dice coefficients of the proposed approach for weighted $\alpha$-softmax function on different $\alpha$'s and $w_{min}$'s. As can be seen, the performance is robust to $w_{min}$ in a large range of $[0.3, 0.7]$, and less robust to $\alpha$. Moreover, the results for the weighted $\alpha$-softmax function were shown in Fig.~\ref{fig:sensitivity_perf}(b), indicating that the performance is robust to both $\alpha$ in $[0.5, 2]$ and $w_{min}$ in $[0.3, 0.7]$.

\subsection{Visualization of the origin in the polar transformation}
\begin{figure}[htbp] 
	\centering
	\setlength{\tabcolsep}{2pt}
	\begin{tabular}{ccc}
	\includegraphics[trim=0in 0in 0in 0in,clip,width=1.1in]{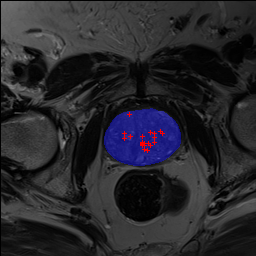} &
	\includegraphics[trim=0in 0in 0in 0in,clip,width=1.1in]{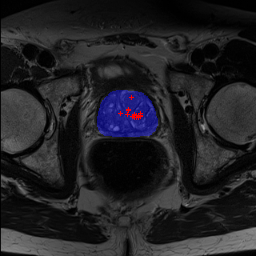} &
	\includegraphics[trim=0in 0in 0in 0in,clip,width=1.1in]{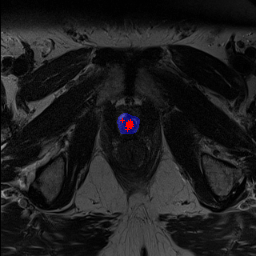} \\ 
	(a) & (b) & (c) \\
	\end{tabular}
	\caption{Selected origins in the polar transformation, where the ground truth of segmentation is marked by blue color and each origin is denoted by a red plus sign.}
	\label{fig:polar_origin}
\end{figure}

To verify the correctness of the selected origin $O$ in the polar transformation, Fig.~\ref{fig:polar_origin} shows selected origins of three examples in the validation subset by the models obtained at the end of each epoch. As can be seen, all origins located in the object, indicating that the proposed approach is able to select origins correctly during training. 

\section{Conclusion}
This study proposed a polar transformation based MIL strategy to assist image segmentation using loose bounding box supervision. The experimental results show that the proposed approach gets superior performance, achieving performance similar to start of the art in the tight bounding box supervision setting.



%
%
%
\bibliographystyle{splncs04}
\bibliography{reference}
\end{document}